\documentclass[iicol]{sn-jnl}

\usepackage{graphicx}%
\usepackage{multirow}%
\usepackage{amsmath,amssymb,amsfonts}%
\usepackage{amsthm}%
\usepackage{mathrsfs}%
\usepackage[title]{appendix}%
\usepackage{xcolor}%
\usepackage{textcomp}%
\usepackage{manyfoot}%
\usepackage{booktabs}%
\usepackage{algorithm}%
\usepackage{algorithmicx}%
\usepackage{algpseudocode}%
\usepackage{listings}%

\usepackage{hyperref}
\usepackage{multirow}
\usepackage{pifont}
\usepackage[authoryear,sort&compress]{natbib}
\begin{document}

\title[NICE: Improving Panoptic Narrative Detection and Segmentation with Cascading Collaborative Learning]{NICE: Improving Panoptic Narrative Detection and Segmentation with Cascading Collaborative Learning}


\author[1]{\fnm{Haowei} \sur{Wang}}
\equalcont{These authors contributed equally to this work.}

\author[1]{\fnm{Jiayi} \sur{Ji}}
\equalcont{These authors contributed equally to this work.}

\author[1]{\fnm{Tianyu} \sur{Guo}}

\author[2]{\fnm{Yilong} \sur{Yang}}
\author[1]{\fnm{Yiyi} \sur{Zhou}}
\author*[1]{\fnm{Xiaoshuai} \sur{Sun}}\email{xssun@xmu.edu.cn}
\author[1]{\fnm{Rongrong} \sur{Ji}}

\affil[1]{\orgdiv{Key Laboratory of Multimedia Trusted Perception and Efficient Computing, Ministry of Education of China}, \orgname{School of Informatics}, \orgaddress{\city{Xiamen}, \country{China}}}


\affil[2]{\orgname{University of Southampton}, \orgaddress{\city{Southampton},  \country{United Kingdom}}}


\abstract{Panoptic Narrative Detection (PND) and Segmentation (PNS) are two challenging tasks that involve identifying and locating multiple targets in an image according to a long narrative description. In this paper, we propose a unified and effective framework called NICE that can jointly learn these two panoptic narrative recognition tasks. Existing visual grounding tasks use a two-branch paradigm, but applying this directly to PND and PNS can result in prediction conflict due to their intrinsic many-to-many alignment property. To address this, we introduce two cascading modules based on the barycenter of the mask, which are Coordinate Guided Aggregation (CGA) and Barycenter Driven Localization (BDL), responsible for segmentation and detection, respectively. By linking PNS and PND in series with the barycenter of segmentation as the anchor, our approach naturally aligns the two tasks and allows them to complement each other for improved performance. Specifically, CGA provides the barycenter as a reference for detection, reducing BDL's reliance on a large number of candidate boxes. BDL leverages its excellent properties to distinguish different instances, which improves the performance of CGA for segmentation. Extensive experiments demonstrate that NICE surpasses all existing methods by a large margin, achieving 4.1\% for PND and 2.9\% for PNS over the state-of-the-art. These results validate the effectiveness of our proposed collaborative learning strategy. The project of this work is made publicly available at \url{https://github.com/Mr-Neko/NICE}.}




\maketitle

\begin{figure}[t]
  \centering
   \includegraphics[width=1\linewidth]{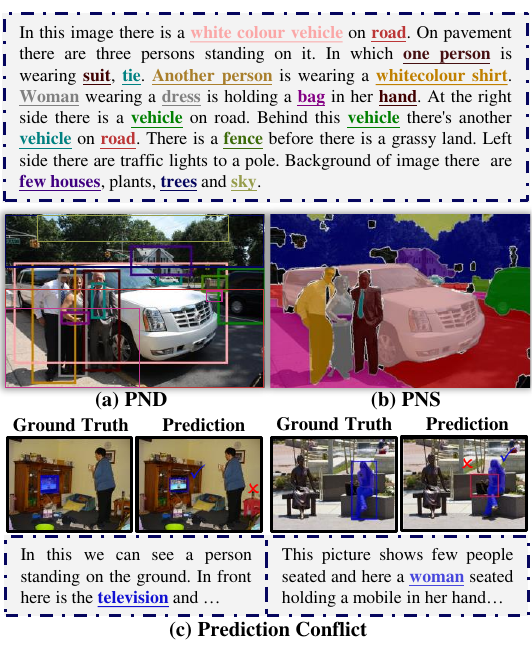}
   \caption{(a) PND locates the target based on the phrase in the description, while (b) PNS completes pixel-level segmentation. (c) The conflicts from two branches pipeline: correct box with wrong mask (left) and wrong box with correct mask (right).}
   \label{fig1}
\end{figure}

\begin{figure}[t]
  \centering
   \includegraphics[width=1\linewidth]{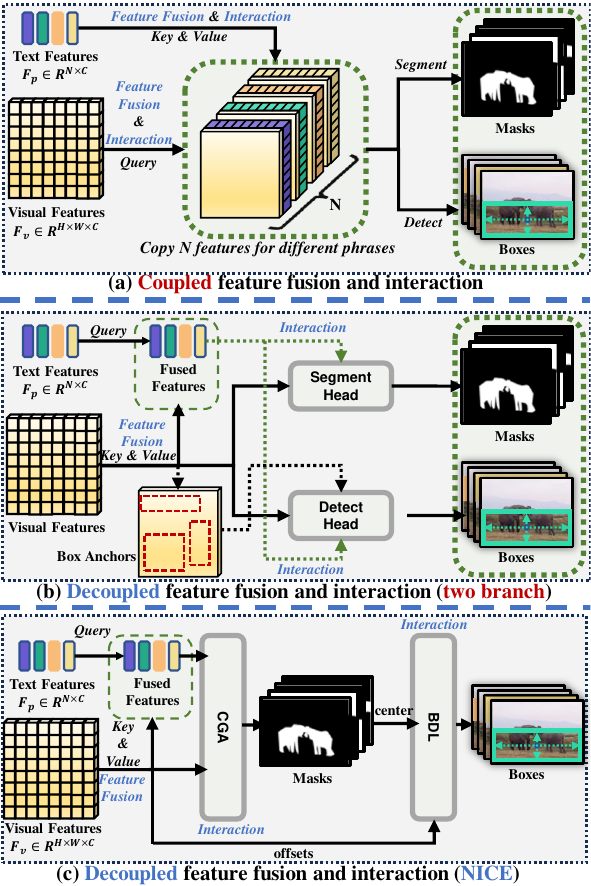}
   \caption{(a) The coupled fusion of features and interaction leads to an increase in computational resources and inference time by $N$ times for PND and PNS. (b) The decoupling of fusion and interaction via a dual-branch approach introduces prediction conflicts, and it concurrently introduces a substantial number of anchors. (c) The NICE framework addresses these issues by simultaneously decoupling the two processes and resolving conflicts using a cascading structure.}
   \label{fig1_p}
\end{figure}

\section{Introduction}
\label{sec:intro}
Panoptic Narrative Detection (PND)~\citep{akbari2019multi, plummer2018conditional, wang2018learning, yu2020cross, mu2021disentangled} and Panoptic Narrative Segmentation (PNS)~\citep{gonzalez2021panoptic, ding2022ppmn} are two novel tasks that necessitate the integration of natural language and visual scenes. While traditional visual grounding tasks such as Referring Expression Comprehension (REC)~\citep{chen2020uniter, jing2020visual, liu2019improving, lu2019vilbert} and Segmentation (RES)~\citep{luo2020multi, karpathy2014deep, ding2021vision} focus on identifying specific objects within an image, PND and PNS require the localization of all objects mentioned in a lengthy narrative description. As depicted in Fig.~\ref{fig1}, this demands a many-to-many alignment between natural language and visual information, making these tasks particularly complex and challenging. Consequently, developing effective approaches to jointly address PND and PNS has become an active research area in computer vision and natural language processing.

Recent studies have treated PND and PNS as distinct tasks that require different approaches~\citep{ding2022ppmn, mu2021disentangled}. Traditionally, PND has been treated as a region-phrase matching problem in which image features are extracted from regions and matched with corresponding features from narrative descriptions using a multi-step pipeline~\citep{nagaraja2016modeling, plummer2018conditional, wang2018learning, kamath2021mdetr}. On the other hand, recent PNS methods have adopted an end-to-end approach to directly match multiple noun phrases with relevant pixels using pixel masks, as opposed to bounding boxes~\citep{ding2022ppmn,wang2023towards}.
Despite the difference in grounding methods, PND and PNS are closely related task definitions except for the grained of the visual target. They can complement each other when modeling natural language grounding in visual scenes. 

To this end, a natural choice is to design a unified multitask framework to accomplish both tasks collaboratively, which, however, is nontrivial due to their many-to-many dense cross-modal alignments. While multi-task learning is prevalent in visual grounding tasks like REC and RES~\citep{luo2020multi, li2021referring}, current methods, which couple interaction and fusion, are not directly translatable to the intricacies of PND and PNS. As illustrated in Fig.~\ref{fig1_p} (a), the prevailing coupled approach requires the replication of visual features multiple times to segment numerous noun phrases, inflating computational and storage costs. Confronted with this inefficiency, our approach seeks to decouple interaction from fusion. Another intuitive strategy, depicted in Fig.~\ref{fig1_p} (b), is to prioritize feature fusion. This then facilitates the direct extraction of both detection and segmentation results during the interaction phase via a separate mechanism. However, striving for precise bounding box predictions, akin to the YOLO strategy~\cite{redmon2016you}, necessitates providing several anchors for each point. This approach substantially reduces inference efficiency and fails to leverage the synergistic potential of PND and PNS. We have observed that point-wise PNS can precisely determine positions. By adopting a strategy of segmentation followed by detection, PND can effectively eliminate its reliance on a large number of anchors and harness the synergistic benefits between the two tasks. 

In this paper, we propose a novel uNIfied Cascading framEwork (NICE) for joint learning of PNG and PND, employing a sequential strategy that detects before segmentingFig.~\ref{fig1_p} (c). To construct the unified framework, our approach, inspired by KNet~\citep{zhang2021k} and PPMN~\citep{ding2022ppmn}, creates a learnable kernel for each noun phrase to predict both its mask and bounding box. To address the challenge of prediction conflict, we introduce two cascade modules, namely the Coordinate Guided Aggregation (CGA) and the Barycenter Driven Localization (BDL), which handle segmentation and detection sequentially while achieving cross-task alignment. We use the barycenter of the segmentation mask as a reference point to ensure natural alignment between the two tasks. Additionally, the position information of the mask drives the BDL to yield accurate bounding boxes. The visual features, that are optimized by the backpropagation of BDL, significantly enhance the ability of CGA to distinguish between different instances during training. 
We retrieve the offset of the barycenter relative to the bounding box using the position information provided by the segmentation and obtain the detection result. This approach eliminates the need to make a separate copy of visual features for each noun phrase, greatly improving the efficiency of PND.

Our contributions are three-fold:

\begin{itemize} 
    \item We present a pioneering collaborative multi-task framework for PND and PNS, which executes segmentation and detection in a sequential, cascading manner.
    \item We introduce two innovative cascading modules, CGA and BDL, which are responsible for segmentation and detection, respectively. These modules are cleverly designed to address the issue of prediction conflict and to facilitate collaborative reasoning.
    \item  Our proposed NICE framework has achieved state-of-the-art performance in both PND and PNS, surpassing previous benchmarks by 2.9\% in segmentation and 4.1\% in detection. Furthermore, NICE exhibits a 22.5\% faster inference speed compared to the single-task network PPMN~\citep{ding2022ppmn}, and reduces memory consumption by a factor of 3 compared to RefTR~\citep{li2021referring}.
\end{itemize}


\section{Related Work}

\subsection{Referring Expression Comprehension and Segmentation}

As prevalent tasks in multi-modal communities, Referring Expression Segmentation (RES)~\citep{hu2016segmentation, ye2019cross, shi2018key, yu2018mattnet, luo2023towards, liu2021cross} and Referring Expression Comprehension (REC)~\citep{nagaraja2016modeling, mao2016generation, hu2016natural, chen2020uniter, chen2022understanding} are to locate one referent based on the understanding of a related short phrase. 

For REC, early models~\citep{hu2017modeling, liu2017referring, luo2017comprehension} obtain a large number of candidate regions in advance through detection models such as RCNN~\citep{girshick2014rich} or YOLO~\citep{redmon2016you, redmon2017yolo9000, redmon2018yolov3}, and then the most suitable box is selected as the output by comparing the features of these regions with the text feature. Recent methods~\citep{sadhu2019zero, yang2016hierarchical} perform the fusion of multimodal features in advance and then perform the detection process, thereby enabling end-to-end training.

For RES, In sequential order, previous models~\citep{li2018referring, liu2017recurrent, margffoy2018dynamic} are like two-stage REC with segmentation models. After that, a batch of methods are developed for refining segmentation masks by a single-stage network~\citep{liu2019learning, zhou2021real}, which bring attracting performance.

\subsection{Panoptic Narrative Detection and Segmentation}

\begin{figure*}[t]
\centering
\includegraphics[width=1.0\textwidth]{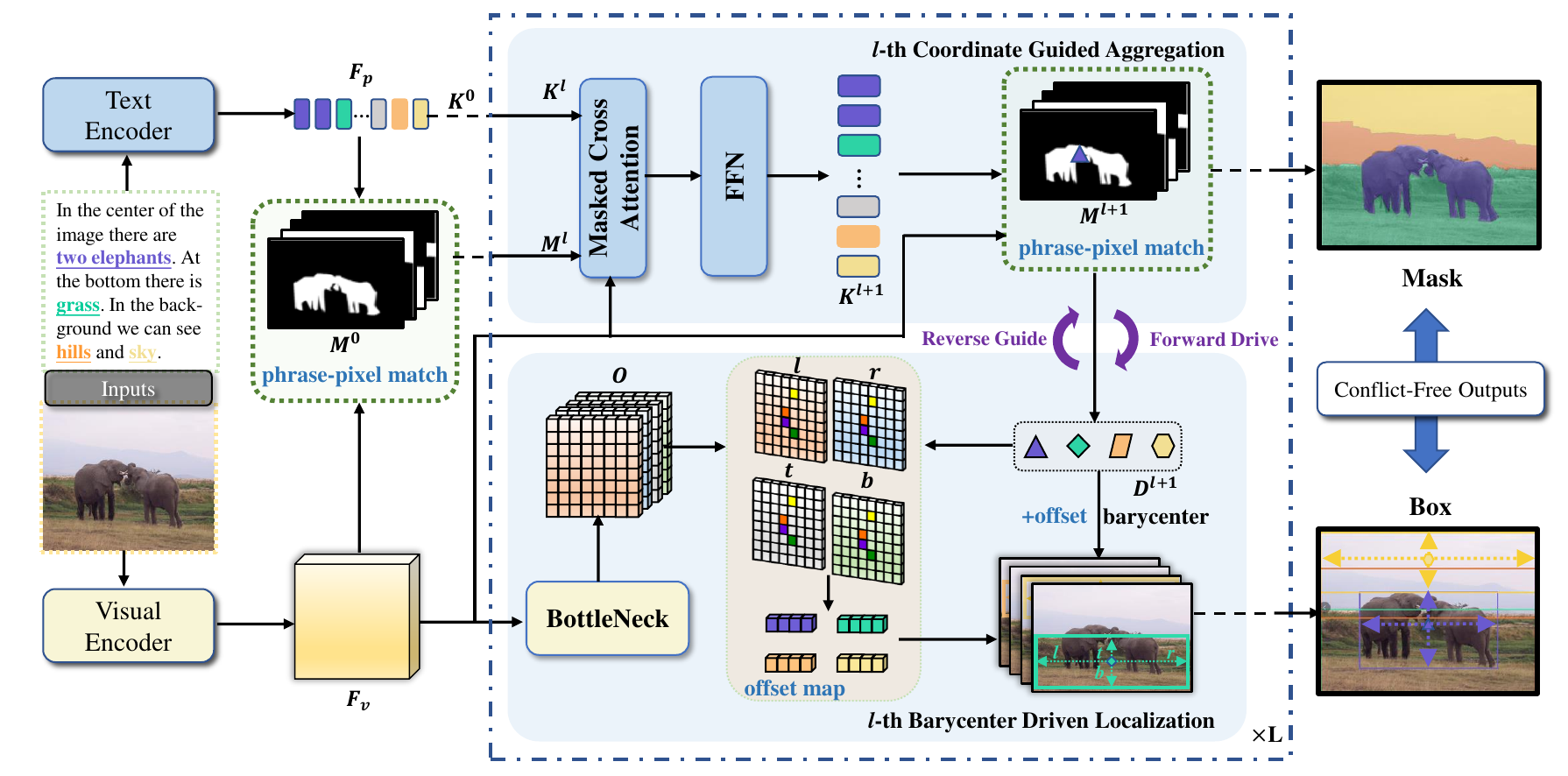} 
\caption{The framework of the proposed NICE. During the pipeline, the text and visual encoders are used to extract the features. Then a cascade framework with L identical layers is used to generate masks and boxes synchronously. In each layer, the kernels interact with visual features and convolve on them to generate masks in CGA. Based on the mask barycenters, BDL predict the final boxes.}
\label{fig2}
\vspace{-0.2em}
\end{figure*}

Panoptic Narrative Detection (PND)~\citep{plummer2018conditional, wang2018learning, bajaj2019g3raphground, karpathy2014deep} and Panoptic Narrative Segmentation (PNS)~\citep{gonzalez2021panoptic, ding2022ppmn, wang2023towards, gonzalez2023piglet, akbari2019multi, DBLP:conf/ijcai/HuiDHWWDHL23} are two tasks to localize multiple panoptic objects, which contain things and stuff, at regional and pixel granularity, respectively, given images and associated narrative text. Different from the above REC and RES tasks that only need to locate a single target, multi-target processing is the key to PND and PNS tasks. 

For PND, also known as phrase localization, earlier methods follow a two-stage approach treat it as a first detection, then matching task~\citep{akbari2019multi, karpathy2014deep, nagaraja2016modeling, plummer2018conditional, wang2018learning}. Such methods often rely on existing object detection methods. After this, end-to-end methods have been proposed, they often rely on feature fusion first and then use the detection head to acquire the target~\citep{luo2020multi}. In addition, there are some methods that rely on graph neural networks to model the relationship between objects and also rely on existing multi-anchor detection paradigms~\citep{bajaj2019g3raphground}.

For PNS, the difference between it and PND is the pixel-level localization task. Gonzalez \emph{et al.}~\citep{gonzalez2021panoptic} first explore the task and propose a two-stage benchmark, which uses the segmentation model~\citep{kirillov2019panoptic} to generate masks, and then perform masks and phrase matching. After that, an end-to-end approach PPMN~\citep{ding2022ppmn} similar to the KNet~\citep{zhang2021k} structure is proposed, which achieved a presentable performance.

\subsection{Multi-task Learning}
The essence of segmentation and detection tasks lies in the understanding of semantic information, so it is a very natural idea to unify multi-task learning. Early methods~\citep{he2017mask} often first obtain the boxes by detection and then segment within the range of the boxes. Such an approach makes the model too dependent on the detection performance. Later, a two-branch end-to-end method appeared, and MCN~\citep{luo2020multi} completed the joint learning of REC and RES in this way. Complex processing is required to align the two tasks in each branch. To this end, we adopt a dynamic kernel manner to build a unified framework to avoid complex processing while reducing the number of anchors.

\section{A Unified Cascading Framework for PND and PNS}

In this section, we give a detailed description of our proposed NICE, of which the framework is illustrated in Fig.~\ref{fig2}. First, the images and descriptions are encoded using separate visual and text encoders. After that, the CGA module facilitates image-text interaction, which enables prediction heads to produce masks. Finally, the BDL module is utilized to generate boxes with the guidance of masks.

\subsection{Feature Extraction}

\subsubsection{Visual Encoder}
Given an image $\mathbf{I} \in \mathbb{R}^{H^0 \times W^0 \times 3}$, we first adopt a FPN~\citep{lin2017feature} with a ResNet-101~\citep{he2016deep} backbone to extract the multi-scale visual features, \emph{e.g.}, $ \mathbf{F_{v}}^{p} \in \mathbb{R} ^ {H^p \times W^p \times C}$, $ p \in \{ 2, 3, 4, 5\} $, where $ H^p = \frac{H^0}{2^p}$ and $ W^p = \frac{W^0}{2^p}$. Considering the importance of position information, we take $F_v^5 $ alone and add it with the position encoding~\citep{ding2022ppmn}. After that, we follow the semantic FPN neck~\citep{kirillov2019panoptic} to stitch them together to get the final visual feature $ \mathbf{F_v} \in \mathbb{R}^{H \times W \times C}$, where $ H = \frac{H^0}{8}$ and $W = \frac{W^0}{8}$.  

\subsubsection{Text Encoder}
Given a sentence $\mathbf{S}$, we follow~\cite{gonzalez2021panoptic} to adopt the pre-trained BERT~\citep{kenton2019bert} to extract the token embeddings $\mathbf{F_{s}}=\{s_t\}_{t=0}^{|S|}$, where $s_t$ denotes the embedding of $t$-th token. Following that, we filter out tokens belonging to noun phrases based on the annotations provided by~\cite{gonzalez2021panoptic, pont2020connecting} and then construct phrase features by average-pooling the token embeddings in each phrase. These phrase features are then projected by a linear layer, aligning their dimension with the visual features, and finally, phrase embeddings $\mathbf{F_{p}} = \{{f_{n}}\}_{n=0}^{N} \in \mathbb{R}^{N \times C}$ are obtained, where $f_n$ represents the feature of the $n$-th noun phrase, and $N$ is the number of phrases. 

Next, both visual features and phrase features are fed into $L$ identical layers, each of which consists of Coordinate Guided Aggregation (CGA) and Barycenter Driven Localization (BDL) in series. Next, we will describe these two modules in detail, taking $\ell$-th layer as an example.

\subsection{Coordinate Guided Aggregation}

Our proposed approach is inspired by KNet~\citep{zhang2021k} and employs an attention map of the interactions between phrase representations and visual features to produce a one-to-one mask for each phrase, thereby eliminating any requirement for matching or post-processing steps. Despite these advances, the similarity between the targets sharing same category poses a challenge in distinguishing them. To address this challenge, we introduce the Coordinate Guided Aggregation (CGA) module, which weakly constrains the segmentation module during training with coordinate information from the bounding boxes.

%
%

More specifically, to further enhance the expressiveness of the text-driven kernels, we first let them interact with visual features, thus aggregating more visual cues. Following~\cite{cheng2022masked}, masked cross-attention is adopted to accomplish it. Using kernels as a query, the attention weights for visual features are calculated as follows:
\begin{equation}
    \mathcal{M}^{\ell-1} = \begin{cases}
    0, &\text{where} \quad \mathbf{M}^{\ell-1} \geq \tau, \\
    {-\infty}, &\text{where} \quad \mathbf{M}^{\ell-1} < \tau,
    \end{cases}
    \label{eq:mask}
\end{equation}
\begin{equation}
        \mathbf{A}^j\!=\!\text{Softmax}\bigg(\frac{(\mathbf{K}^{\ell-1}\mathbf{W}^j_Q)(\mathbf{F}_{v}\mathbf{W}^j_K)^T}{\sqrt{d_k}}\!+\!\mathcal{M}^\mathbf{\ell-1} \bigg),
    \label{eq:Attention}
\end{equation}
where $\tau$ is the threshold to control the $\mathcal{M}^\mathbf{\ell-1}$, $\mathbf{M}^{\ell-1} \in [0, 1]^{N \times H \times W}$ represents the masks of the $(\ell-1)$-th layer.  The projections $\mathbf{W}^j_Q \in \mathbb{R}^{C \times \frac{C}{h}}$ and $\mathbf{W}^j_K \in \mathbb{R}^{C \times \frac{C}{h}}$ are weight matrices, and $d_k$ is a scaling factor. The subscript $j$ represents the $j$-th head and the number of heads $h$ is set to 8. Specially, $\mathbf{K}^0$ is obtained after the initialization of phrase embeddings $\mathbf{F_p}$, and $\mathbf{M^0}$ is given by:
\begin{equation}
    \mathbf{M}^0 = \text{Sigmoid}\left ( \mathbf{K}^0 \ast \mathbf{F_v} \right ).
    \label{eq:0}
\end{equation}

$\mathcal{M}^\mathbf{\ell-1}$ limits the scope of the interaction of the kernel with visual features, thus avoiding the inclusion of irrelevant semantics. Based on these attention weights, we aggregate semantically related visual features for head $j$ and concatenate all the results to get the updated kernels:
\begin{equation}
    \text{Head}^j = \mathbf{A}^j(\mathbf{K}^{\ell-1}\mathbf{W}^j_V),
    \label{eq:vi}
\end{equation}
\begin{small}
\begin{equation}
    \mathbf{K}^{\ell} = \text{FFN}\left(\text{LN}\left(\left[\text{Head}^1, \cdots, \text{Head}^h\right]\mathbf{W}_O\right) + \mathbf{K}^{\ell-1}\right),
    \label{eq:MHA}
\end{equation}
\end{small}
where $\mathbf{W}^j_V\in \mathbb{R}^{C \times \frac{C}{h}}$, and $\mathbf{W}_O\in \mathbb{R}^{C \times C}$ are projection matrices. Besides, $\text{LN}\left(\cdot\right)$ means the \emph{layer normalization}~\citep{ba2016layer}. With shortcut connection~\citep{he2016deep} is applied after the $\text{LN}\left(\cdot\right)$, a $\text{FFN}\left(\cdot\right)$, \emph{Feed-Forward Network}~\citep{vaswani2017attention} is used to map features.

Finally, these text-driven dynamic kernels will operate on the visual features to get the final masks:
\begin{equation}
    \mathbf{M}^\ell = \text{Sigmoid}\left ( \mathbf{K}^\ell \ast \mathbf{F_v} \right ).
    \label{eq:1}
\end{equation}
The text-driven kernel is continuously updated at each layer to generate a more accurate mask. as the layers increase, both are optimized to promote each other. 

In addition, we further constrain the segmentation with the help of detection boxes to assist in distinguishing different instances. One of the most straightforward solutions is to find the smallest rectangular box that can enclose the mask as the detection box. In this way, detection and segmentation can be done in one step. However, such a restriction is too strict, resulting in a detection performance that depends entirely on the segmentation boundary. The segmentation boundary is extremely difficult to handle, which greatly limits the performance of the detection. For this purpose, we decouple the boundaries by using a soft constraint, i. e., the center of the mask, to connect the two modules in series, as described in the next subsection.

\subsection{Barycenter Driven Localization}
\label{sec:BDL}

%
In previous multi-task learning approaches, segmentation and detection usually flow into two parallel branches, so additional modules are needed to enhance the synergy of these two tasks. As we stated in the previous subsection, the mask generated by the CGA can roughly mark the location of the target, which can be regarded as a very practical cue for PND. Therefore, we propose a new module, \emph{i.e.}, BDL, which is plugged behind the CGA, so that BDL can take full advantage of the general location information CGA provides. 

Earlier methods made a copy of the visual features for each noun phrase and migrated directly to a setting such as PND where many nouns are generated at once, making detection much less efficient. We observe that once the position is determined, if there are targets at that point, their size is determined. Thus we use the visual feature $\mathbf{F_v}$ to predict the offset value of each predicted box relative to the barycenters by:
\begin{equation}
    \mathbf{O} = \text{Sigmoid}\left(\text{BottleNeck}\left(\mathbf{F_v}\right)\right),
\label{eq:offset}
\end{equation}
where $\text{BottleNeck}\left(\cdot\right)$ is a cascaded convolution network~\citep{tian2019fcos} to change the dimensions of feature and $\mathbf{O} \in [0, 1]^{H \times W \times 4}$ is the offset matrix. In this way, only one visual feature is used to predict bounding boxes at a time, significantly reducing time and memory overhead.

In general, the center of the mask is also, with high probability, the center of the target entity. Therefore, we need to find the barycenters of the mask first. With the masks $\mathbf{M^\ell}$ from the $\ell$-th CGA layer, the barycenters are given by:
\begin{equation}
    \mathbf{D_{x}}^{\ell, n} = \dfrac{\iint_{M^\ell} x^n\mathbf{M^\ell}\left(x^n, y^n\right) dxdy}{\sum_{x^n, y^n}^{W, H} \mathbf{M^\ell}\left(x^n, y^n\right)},
\end{equation}
\begin{equation}
    \mathbf{D_{y}}^{\ell, n} = \dfrac{\iint_{M^\ell} y^n\mathbf{M^\ell}\left(x^n, y^n\right) dxdy}{\sum_{x^n, y^n}^{W, H} \mathbf{M^\ell}\left(x^n, y^n\right)},
\end{equation}
where $n$ means the $n$-th noun phrase, and $\mathbf{D}^\ell \in \mathbb{R}^{N \times 2}$ is the two-dimensional coordinates of the mask barycenters.

Finally, we predict bounding boxes for all the noun phrases by combining the barycenter and its corresponding offset as:
\begin{equation}
    \begin{cases}
     x^{1} = \mathbf{D_{x}}^{\ell, n} - l, \\
     y^{1} = \mathbf{D_{y}}^{\ell, n} - t, \\
     x^{2} = \mathbf{D_{x}}^{\ell, n} + r, \\
     y^{2} = \mathbf{D_{y}}^{\ell, n} + b, 
    \end{cases}
\end{equation}
where $l, t, r, b$ are from the $\mathbf{O}_{x, y}$ in eq.~\ref{eq:offset}, which means the distance between the barycenter and the edges of the referred bounding box. $\mathbf{D_{x}}^{\ell, n}$ and $\mathbf{D_{y}}^{\ell, n}$ mean the coordinate of the barycenter, and the $x^1$ and $y^1$ mean the coordinate of the left-top point of the bounding box, while $x^2$ and $y^2$ are the right-bottom point . 

With the general location information provided by the mask, BDL can maintain a high accuracy without a large number of candidate frames. At the same time, such a link facilitates the synergy between the two tasks and enhances their performance.

\subsection{Training Loss}
Since a unified framework for PND and PNS is involved, we require both segmentation and detection objectives to optimize the model. Next, we elaborate on them in detail.

\subsubsection{Segmentation Loss}
With the ground truth mask $\mathbf{Y} \in \mathbb{R}^{N \times H \times W}$ and the prediction $\mathbf{M} \in \mathbb{R}^{N \times H \times W}$, we take BCE loss~\citep{milletari2016v} $\mathcal{L}_{bce}$ and Dice loss~\citep{milletari2016v} to optimizer the effect of segmentation. 
For the whole phrases, we can rewrite BCE loss as:
\begin{equation}
    \overline{\mathcal{L}}_{bce} = -\frac{1}{H \times W} \sum_{i=0}^{H \times W} \frac{1}{N} \sum_{n=0}^{N} \mathcal{L}_{bce}\left(\mathbf{M}^{n, i}, \mathbf{Y}^{n, i}\right).
\end{equation}
Similarly, the Dice loss is:

\begin{equation}
    \overline{\mathcal{L}}_{dice} = \frac{1}{N} \sum_{n=0}^{N} 1-\frac{2|\mathbf{M}^n\bigcap \mathbf{Y}^n|}{|\mathbf{M}^n|+|\mathbf{Y}^n|}.
\end{equation}

\subsubsection{Detection Loss}
As for the detection, we use the Smooth L1 loss~\citep{girshick2015fast} and gIoU loss~\citep{rezatofighi2019generalized} to constrain the prediction boxes. The Smooth L1 loss is:

\begin{small}
\begin{equation}
    \mathcal{L}_{s}\left( \mathbf{B}^{n, i}, \mathbf{G}^{n, i}\right) \!=\!\begin{cases}
    0.5\left(\mathbf{B}^{n, i}\!-\!\mathbf{G}^{n, i}\right)^2, &|x|\!<\!\xi \\
    |\mathbf{B}^{n, i}\!-\!\mathbf{G}^{n, i}|\!-\!0.5, &otherwise,
    \end{cases}
    \label{loss:l1}
\end{equation}
\end{small}

\noindent where $x = \mathbf{B}^{n, i}-\mathbf{G}^{n, i}$, $\mathbf{B}^n \in \mathbb{R}^{4}$ means the prediction box of the $n$-th phrase, the $\mathbf{G}^n$ is the correspoding ground truth, and $ i \in \{x, y, w, h\}$. We set the $\xi$ as 0.5 during training.
For all the phrase, Eq.~\ref{loss:l1} can be rewritten as:
\begin{equation}
    \overline{\mathcal{L}}_{smooth} = \frac{1}{4N}\sum_{n=0}^{N}\sum_{i}\mathcal{L}_{s}\left( \mathbf{B}^{n, i}, \mathbf{G}^{n, i}\right).
\end{equation}

Since the optimization of Smooth L1 loss is not equivalent to the optimization of IoU, we also add the gIoU loss, which is:
\begin{equation}
    \overline{\mathcal{L}}_{gIoU} = \frac{1}{N} \sum_{n=0}^{N} 1-\frac{\mathbf{B}^n\bigcap \mathbf{G}^n}{\mathbf{B}^n \bigcup \mathbf{G}^n}+\frac{A_c^n - \mathbf{B}^n \bigcup \mathbf{G}^n}{A_c^n},
\end{equation}
where $A_c^n$ is the minimum closure area of $\mathbf{B}^n$ and $\mathbf{G}^n$.

In summary, the final loss is the weighted combination of the four losses above:
\begin{equation}
    L = \lambda_{1}\overline{\mathcal{L}}_{bce} + \lambda_{2}\overline{\mathcal{L}}_{dice} + \lambda_{3}\overline{\mathcal{L}}_{smooth} + 
    \lambda_{4}\overline{\mathcal{L}}_{gIoU}.
\end{equation}

\begin{table*}[t!]
\centering
\resizebox{0.96\linewidth}{!}{
\begin{tabular}{l|cccccc|cccccc|c|c|c}
\toprule
\multirow{2}{*}{Method} & \multicolumn{6}{|c|}{Segmentation Average Recall} & \multicolumn{6}{|c|}{Detection Average Recall} & \multirow{2}{*}{IE $\downarrow$} & \multirow{2}{*}{Gflops} & \multirow{2}{*}{Params} \\

\cmidrule{2-13}
 & All & Thing & Stuff & Single & Plural & Time & All & Thing & Stuff & Single & Plural & Time & & & \\

\hline
PNG~\citep{gonzalez2021panoptic} & 55.4 & 56.2 & 54.3 & 56.2 & 48.8 & 107ms & -& -& -& -& - &- &- & \textbf{62.5} & 100.1 M \\
EPNG~\citep{wang2023towards} & 58.0 & 54.8 & 62.4 & 58.6 & 52.1 & \textbf{35.5ms} & -& -& -& -& - &- &- & 259.5 & 24.2 M \\
PPMN~\citep{ding2022ppmn} & 59.4 & 57.2 & 62.5 & 60.0 & 50.4 & 63.5ms & -& -& -& -& - &- &- & 241.1 & 10 M\\
RefTR~\citep{li2021referring}  & - & - & - & - & - & - & 53.6 & 47.2 & 62.4 & 54.0 & 50.2 & 82.1ms & - & 808.1 & 12.3 M\\
MCN~\citep{luo2020multi} & 54.2 & 48.6 & 61.4 & 56.6 & 38.8 & 146.3ms & 49.9 & 49.8 & 50.0 & 50.7 & 44.7& 146.3ms & 20.3\% & 342.8 & 14.5 M\\

\hline

NICE$^{\dagger}$ & 58.6 & 56.0 & 62.2 & 59.5 & 50.1 & 49.2ms & 45.4 & 37.8 & 55.8 & 45.6 & 43.7 & \textbf{49.2ms}& 27.2\% & 164.0 & 8.6 M\\
NICE$^{*}$ & 60.8 & 58.7 & 63.8 & 61.7 & 52.8 & 49.2ms & 47.5 & 39.3 & 58.8 & 47.8 & 44.7 & 53.6ms & 30.5\% & 164.0 & \textbf{8.6 M}\\
NICE & \textbf{62.3} & \textbf{60.2} & \textbf{65.3} & \textbf{63.1} & \textbf{55.2} & 49.2ms & \textbf{57.7} & \textbf{54.1} & \textbf{62.7} & \textbf{58.7} & \textbf{49.2} & 56.9 ms & \textbf{18.3\%} & 285.2 & 10.9 M\\

 \bottomrule
\end{tabular}
}
\caption{Comparison of the NICE and the state-of-the-art methods. MCN is a mluti-task learning method with a two-branch structure. NICE$^{\dagger}$ is the variant using the two-branch design, just like MCN. NICE$^{*}$ directly finds the smallest rectangular box that can enclose the mask as the detection box.}
\label{table1}
\end{table*}

\begin{figure*}[t]
  \centering
   \includegraphics[width=1\textwidth]{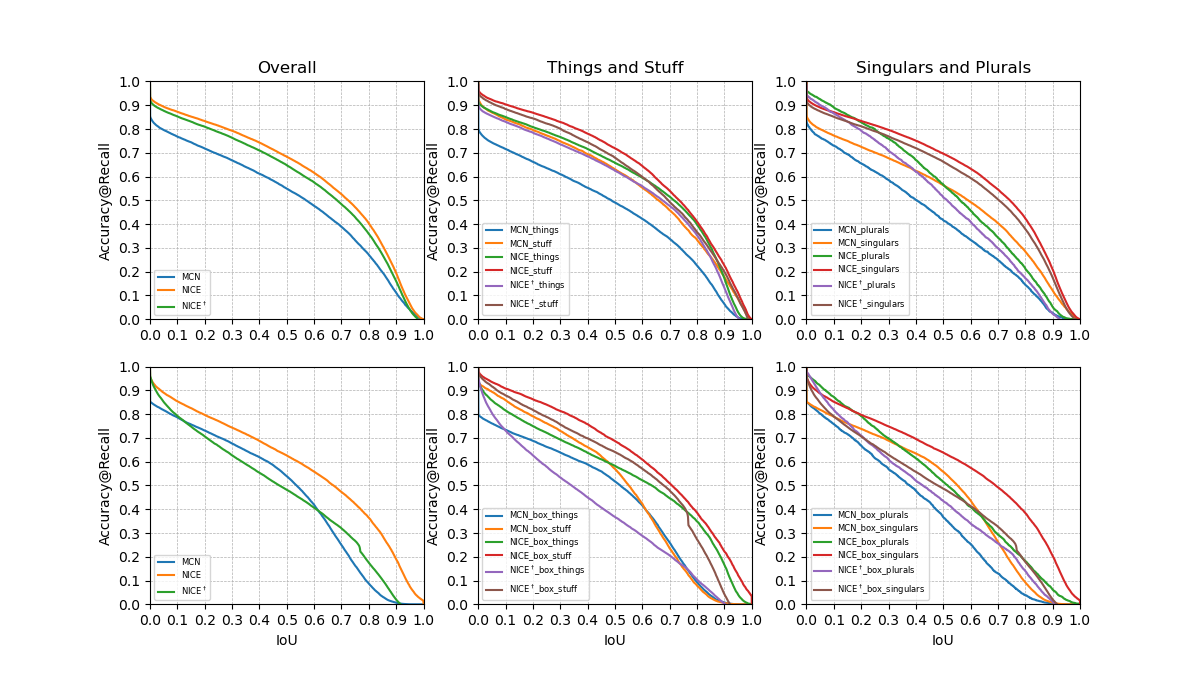}
   \caption{The curve of Average Recall from MCN, NICE$^\dagger$, and NICE. The top line is the recall of PNS and the bottom line is the counterpart of PND.}
   \label{sm_fig1}
\end{figure*}

\begin{table*}[]
\centering
{
\begin{tabular}{l|ccccc|ccccc}
\toprule
\multirow{2}{*}{Method} & \multicolumn{5}{|c|}{Segmentation Average Recall} & \multicolumn{5}{|c}{Detection Average Recall}\\

\cmidrule{2-11}
 & All & Thing & Stuff & Single & Plural & All & Thing & Stuff & Single & Plural\\

\hline
OnlyBox & - & - & - & - & - & 43.2 & 37.4 & 51.2 & 43.3 & 42.2\\
OnlyMask & 58.1 & 54.6 & 63.0 & 59.3 & 47.9 & -& -& -& -& -\\
NICE &\textbf{62.3} & \textbf{60.2} & \textbf{65.3} & \textbf{63.1} & \textbf{55.2} & \textbf{57.7} & \textbf{54.1} & \textbf{62.7} & \textbf{58.7} & \textbf{49.2}\\

 \bottomrule
\end{tabular}}
\caption{Ablation study of single and multi-task learning. OnlyBox and OnlyMask are single-task learning.}
\label{table2}
\vspace{-0.5em}
\end{table*}

\begin{table*}[]
\centering
{
\begin{tabular}{c|c|ccccc|ccccc}
\toprule
\multicolumn{2}{c|}{Method} & \multicolumn{5}{|c|}{Segmentation Average Recall} & \multicolumn{5}{|c}{Detection Average Recall}\\

\hline
 CGA & BDL & All & Thing & Stuff & Single & Plural & All & Thing & Stuff & Single & Plural\\

\hline
\ding{55} & \ding{55} & 58.6 & 56.0 & 62.2 & 59.5 & 50.1 & 45.4 & 37.8 & 55.8 & 45.6 & 43.7 \\
\ding{51} & \ding{55} & 60.4 & 57.7 & 64.0 & 61.2 & 52.6 & 45.9 & 33 & \textbf{63.9} & 46.3 & 42.9\\
\ding{51} & \ding{51} & \textbf{62.3} & \textbf{60.2} & \textbf{65.3} & \textbf{63.1} & \textbf{55.2} & \textbf{57.7} & \textbf{54.1} & 62.7 & \textbf{58.7} & \textbf{49.2}\\

 \bottomrule
\end{tabular}}
\caption{Ablation study of CGA and BDL.}
\label{table3}
\end{table*}

\begin{table*}[]
\centering
{
\begin{tabular}{l|ccccc|ccccc}
\toprule
\multirow{2}{*}{Barycenter} & \multicolumn{5}{|c|}{Segmentation Average Recall} & \multicolumn{5}{|c}{Detection Average Recall}\\

\cmidrule{2-11}
 & All & Thing & Stuff & Single & Plural & All & Thing & Stuff & Single & Plural\\

\hline
Top1 & 59.6 & 56.5 & 63.8 & 60.4 & 51.9 & 30.4 & 22.9 & 40.9 & 30.4 & 30.3\\
Top50 & 61.1 & 58.7 & 64.5 & 62 & 52.6 & 41.1 & 37.2 & 46.6 & 41.9 & 33.9\\
Top100 & 59.7 & 57.1 & 63.3 & 60.7 & 50.6 & 35.1 & 31.2 & 40.5 & 35.6 & 30.0\\
Average &\textbf{62.3} & \textbf{60.2} & \textbf{65.3} & \textbf{63.1} & \textbf{55.2} & \textbf{57.7} & \textbf{54.1} & \textbf{62.7} & \textbf{58.7} & \textbf{49.2}\\

 \bottomrule
\end{tabular}}
\caption{Ablation study of the settings of barycenters.}
\label{table4}
\end{table*}

\begin{table*}[]
\centering
\resizebox{!}{!}{
\begin{tabular}{l|ccccc|ccccc}
\toprule
\multirow{2}{*}{Scale} & \multicolumn{5}{|c|}{Segmentation Average Recall} & \multicolumn{5}{|c}{Detection Average Recall}\\

\cmidrule{2-11}
 & All & Thing & Stuff & Single & Plural & All & Thing & Stuff & Single & Plural\\

\hline
Mluti-scale & 60.5 & 58.1 & 63.8 & 61.4 & 52.3 & \textbf{58.7} & 54.0 & \textbf{65.2} & \textbf{59.7} & \textbf{49.7}\\
Fusion &\textbf{62.3} & \textbf{60.2} & \textbf{65.3} & \textbf{63.1} & \textbf{55.2} & 57.7 & \textbf{54.1} & 62.7 & 58.7 & 49.2\\

 \bottomrule
\end{tabular}}
\caption{Ablation study of scales in BDL.}
\label{table5}
\vspace{-0.5em}
\end{table*}

\begin{table*}[]
\centering
{
\begin{tabular}{l|ccccc|ccccc}
\toprule
\multirow{2}{*}{Method} & \multicolumn{5}{|c|}{Segmentation Average Recall} & \multicolumn{5}{|c}{Detection Average Recall}\\

\cmidrule{2-11}
 & All & Thing & Stuff & Single & Plural & All & Thing & Stuff & Single & Plural\\

\hline
w/o stuff & 59.0 & 56.1 & 63.2 & 59.7 & 53.2 & 48.9 & 48.9 & -& 46.4& 46.7\\

NICE &\textbf{62.3} & \textbf{60.2} & \textbf{65.3} & \textbf{63.1} & \textbf{55.2} & \textbf{57.7} & \textbf{54.1} & \textbf{62.7} & \textbf{58.7} & \textbf{49.2}\\

 \bottomrule
\end{tabular}}
\caption{Ablation study of with and without stuff. }
\label{tab: stuff}
\end{table*}

\begin{table*}[t]
\centering
\begin{tabular}{c|c|c|ccccc}
\toprule
 \multicolumn{2}{c|}{Dataset} & Type & AP@0.1 & AP@0.2 & AP@0.3 & AP@0.4 & AP@0.5\\
 \hline
 \multirow{4}{*}{RefCOCO} & \multirow{2}{*}{testA} & random & 36.2 & 10.4 & 3.3 & 1.1 & 0.2 \\
 \cmidrule{3-3}
  & & zero-shot & \textbf{51.0} & \textbf{47.4} & \textbf{43.4} & \textbf{39.3} & \textbf{34.8}\\
  \cmidrule{2-8}
  & \multirow{2}{*}{testB} & random & 40.1 & 13.2 & 5.0 & 2.3 & 0.6 \\
 \cmidrule{3-3}
  & & zero-shot & \textbf{60.6} & \textbf{53.0} & \textbf{45.9} & \textbf{37.8} & \textbf{30.0}\\
  
  \hline
 \multirow{4}{*}{RefCOCO+} & \multirow{2}{*}{testA} & random & 34.1 & 9.7 & 3.0 & 1.1 & 0.1 \\
 \cmidrule{3-3}
  & & zero-shot & \textbf{34.1} & \textbf{29.9} & \textbf{26.8} & \textbf{23.4} & \textbf{19.6}\\
  \cmidrule{2-8}
  & \multirow{2}{*}{testB} & random & 40.3 & 12.9 & 5.1  & 2.5 & 0.6 \\
 \cmidrule{3-3}
  & & zero-shot & \textbf{44.0} &\textbf{ 37.0} & \textbf{30.9} & \textbf{24.5} & \textbf{18.8}\\
  
\hline
 \multirow{2}{*}{RefCOCOg} & \multirow{2}{*}{test} & random & 39.0 & 14.9 & 5.8 & 2.3 & 1.0 \\
 \cmidrule{3-3}
  & & zero-shot & \textbf{46.6} &\textbf{ 41.2} & \textbf{36.6} & \textbf{31.4} & \textbf{26.2}\\

 \bottomrule
\end{tabular}
\caption{Zero-shot results of NICE on RES. We cacluate the precious when the threshold of IoU is from 0.1 to 0.5.}
\label{tablezs}
\end{table*}

\begin{table*}[t]
\centering
\begin{tabular}{c|c|c|ccccc}
\toprule
 \multicolumn{2}{c|}{Dataset} & Type & AP@0.1 & AP@0.2 & AP@0.3 & AP@0.4 & AP@0.5\\
 \hline
 \multirow{4}{*}{RefCOCO} & \multirow{2}{*}{testA} & random & 66.7 & 34.1 & 17.4 & 9.0 & 4.2 \\
 \cmidrule{3-3}
  & & zero-shot & \textbf{67.4} &\textbf{ 57.4} & \textbf{48.5} & \textbf{39.3} & \textbf{32.3}\\
  \cmidrule{2-8}
  & \multirow{2}{*}{testB} & random & 67.9 & 33.8 & 17.6 & 8.8 & 5.0 \\
 \cmidrule{3-3}
  & & zero-shot &\textbf{ 77.6} & \textbf{61.2} & \textbf{47.6} & \textbf{35.8} & \textbf{27.4}\\
  
  \hline
 \multirow{4}{*}{RefCOCO+} & \multirow{2}{*}{testA} & random & 53.8 & 32.1 & 16.4 & 8.4 & 3.8 \\
 \cmidrule{3-3}
  & & zero-shot & \textbf{54.8} & \textbf{43.0} & \textbf{33.2} & \textbf{24.8} & \textbf{18.6}\\
  \cmidrule{2-8}
  & \multirow{2}{*}{testB} & random & 67.4 & 34.1 & 17.5 & 8.9 & 5.1 \\
 \cmidrule{3-3}
  & & zero-shot & \textbf{68.4} & \textbf{50.8} & \textbf{37.8} &\textbf{ 26.1} & \textbf{18.5}\\
  
\hline
 \multirow{2}{*}{RefCOCOg} & \multirow{2}{*}{test} & random & 67.6 & 36.4 & 21.2 & 12.8 & 7.4 \\
 \cmidrule{3-3}
  & & zero-shot & \textbf{80.0} & \textbf{63.4} & \textbf{47.4} & \textbf{34.0} & \textbf{23.5}\\
 \bottomrule
\end{tabular}
\caption{Zero-shot results of NICE on REC. We cacluate the precious when the threshold of IoU is from 0.1 to 0.5.}
\label{tablezb}
\end{table*}

\begin{table}[]
    \centering
    \resizebox{1.0\columnwidth}{!}{
    
    \begin{tabular}{c|c|c|c}
        \toprule
       Method  & Proposals & AP50 & Inference Time (ms)\\
       \midrule
       CITE~\citep{plummer2018conditional} & 200 & 61.33 & 196 \\ 
       FAOA~\citep{yang2019fast} & 0 & 67.08 & 38 \\
       NICE & 0 & \textbf{68.71} & \textbf{11} \\
       \bottomrule
    \end{tabular}}
    \caption{Visual grounding on Flickr30K.}
    \label{tab:flickr30k}
\end{table}


\begin{figure*}[t]
\centering
\includegraphics[width=0.96\textwidth]{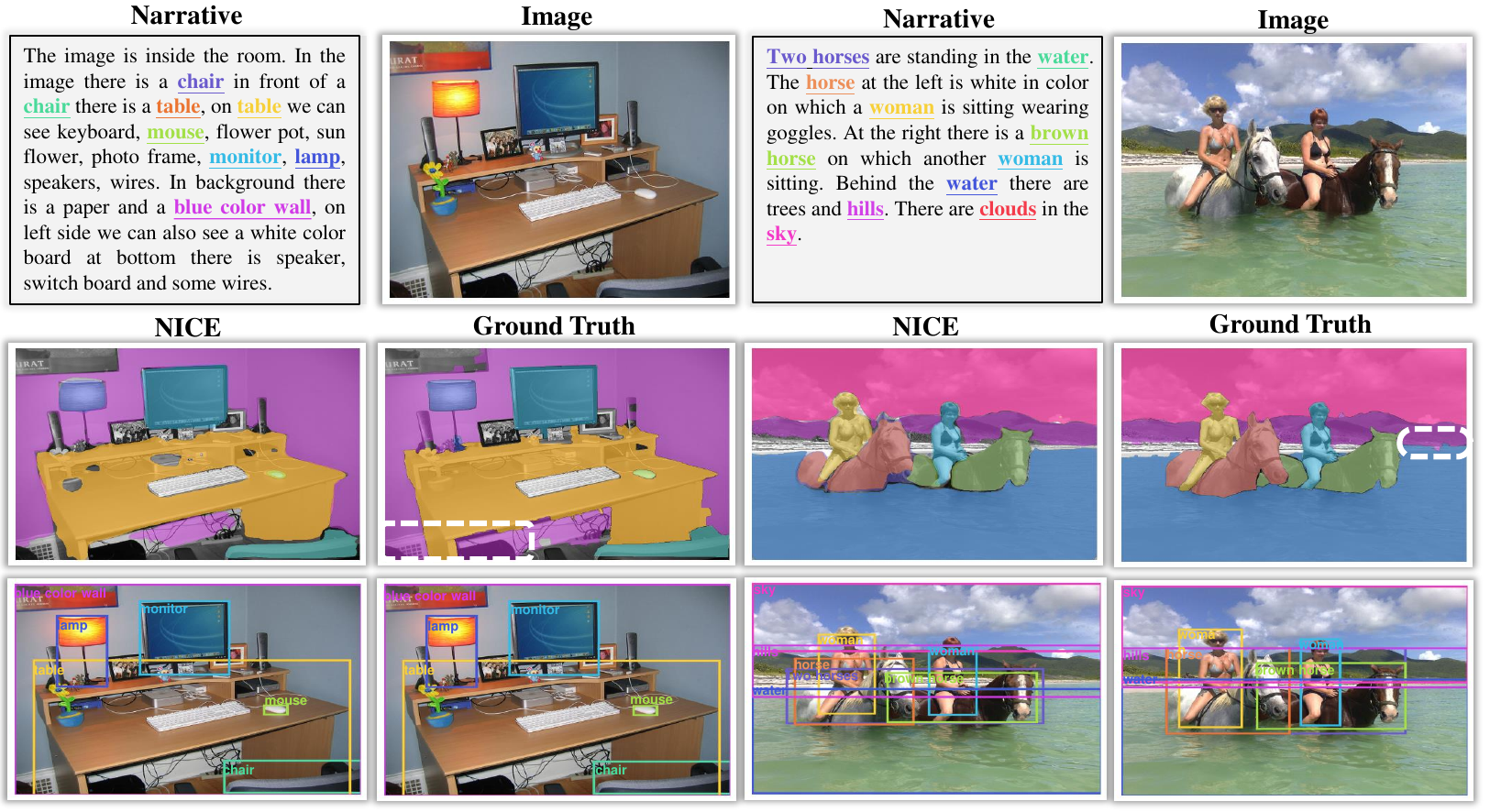} 
\caption{Visualizations of NICE's predictions. We use the same color to mark the masks and boxes with their corresponding phrases. In particular, we highlight the regions, where NICE does better, with white dashed boxes on the ground truth.}
\label{fig4}
\vspace{-1em}
\end{figure*}

\begin{figure*}[t]
\centering
\includegraphics[width=1.0\textwidth]{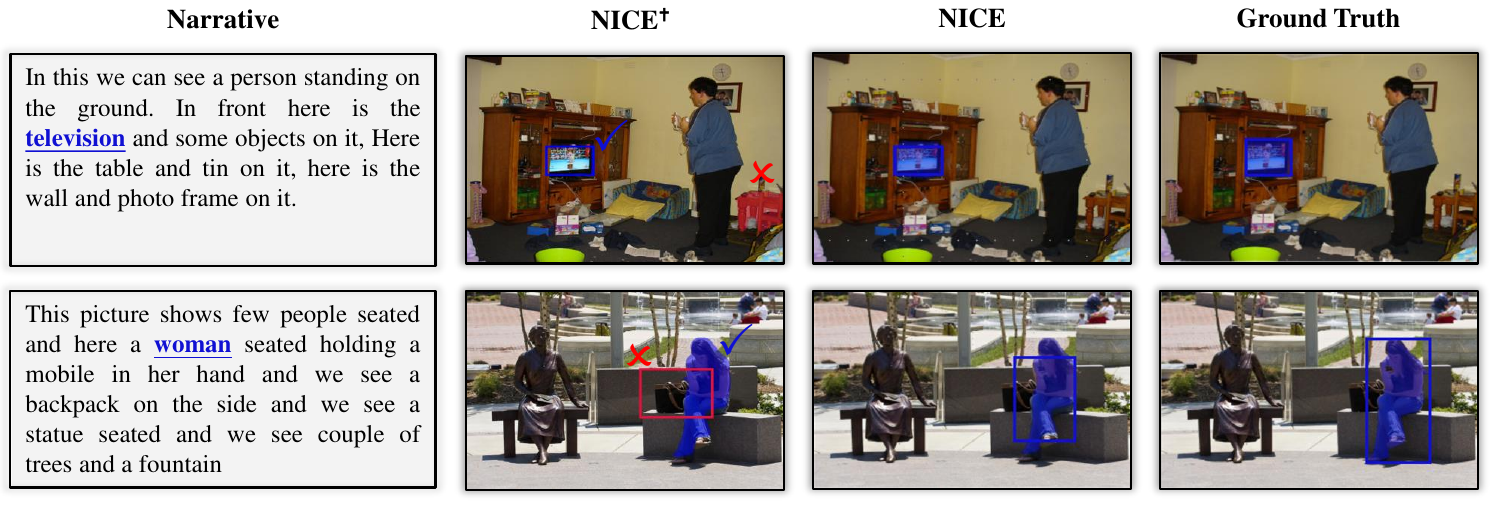} 
\vspace{-0.9cm}
\caption{The comparison between NICE and NICE$^\dagger$. NICE solves the prediction conflict with the proposed CGA and BDL.}
\label{fig5}
\end{figure*}

\section{Experiment}

\subsection{Datasets}
To validate the effectiveness of NICE in addressing PND and PNS, we conducted a study using the Panoptic Narrative Grounding (PNG) dataset~\citep{gonzalez2021panoptic} and compared it with existing methods. The PNG dataset is unique because it includes both detection and segmentation annotation of phrase content in image-text pairs. Unlike datasets such as RefCOCO~\citep{yu2016modeling}, ReferIt~\citep{kazemzadeh2014referitgame}, or RefCOCOg~\citep{mao2016generation, nagaraja2016modeling}, where each brief phrase corresponds to only one target, the PNG dataset contains an average of 5.1 objects per long narrative, including both stuff and things. The narratives are typically several hundred words long and contain more complex semantic content. 
The dataset comprises 133,103 training images and 8,380 test images, with each image containing 875,073 mask and box annotations.

\subsection{Implementation Details}
\subsubsection{Experimental Settings }
To develop our model, we utilize ResNet-101~\citep{he2016deep} as our visual backbone, which is pre-trained on COCO~\citep{lin2014microsoft}. For the text backbone, we adopt BERT following~\citep{gonzalez2021panoptic}. During the training phase, the visual backbone is fixed, and we raise the input image resolution to $640 \times 640$ for parallel training, resulting in fused visual features of $80 \times 80 \times 256$ dimensions. The dimension of the text features is set at 768, with a hidden dimension of 2048 and 8 attention heads. We use 3 layers for our model. In terms of hyperparameters, we balance the final loss using $\lambda_{1}=1$, $\lambda_{2}=1$, $\lambda_{3}=1$, and $\lambda_{4}=1$. We choose the masks from all layers and the boxes from the last layer to participate in the training process. The initial learning rate was set at $\eta=1e^{-4}$, which is reduced by 50\% every 5 epochs and fixed at $\eta=5e^{-7}$ after 10 epochs. The batch size is 48, and we train the model on four RTX3090 GPUs for a total of 20 hours. We utilize the Adam optimizer during training.

\subsubsection{Metrics}
To keep PNS and PND consistent, in this paper, we adopt Average Recall~\citep{gonzalez2021panoptic} as our metric for both segmentation and detection. Specifically, for all the noun phrases from the description, after obtaining the predicted masks and boxes through the model, we will calculate the IoU~\citep{jiang2018acquisition} with the corresponding ground truths, respectively. After obtaining the IoU of all targets, we set different thresholds to calculate the corresponding recall, and calculate the integral of this recall curve as the Average Recall. For boxes and masks, we compute Average Recall on all phrases, things, stuff, singulars, and plurals simultaneously to obtain a comprehensive and detailed model evaluation.

\subsection{Comparison with State-of-the-Art Methods}

We conducted comprehensive experiments on the PNG dataset to evaluate the performance of our proposed NICE model and compare it with state-of-the-art methods. Our results, summarized in Tab.~\ref{table1}, show that NICE outperforms the two-stage pipeline PNG~\citep{gonzalez2021panoptic} and the one-stage pipeline PPMN~\citep{ding2022ppmn} on the PNS task, which is known to have the best performance in recent works. We also introduced MCN~\citep{luo2020multi} to validate our model on multi-task learning, which jointly learns the REC and RES.

Our NICE model achieved an impressive performance boost of 6.9\%/4.0\%/11.0\%/6.9\%/6.4\% on the Average Recall of /all/thing/stuff/single/plural splits with $2\times$ the inference speed compared to PNG (107ms \emph{vs.} 49.2ms). Similarly, our model outperformed PPMN in terms of all metrics with faster inference, demonstrating its superiority in the PNS task and setting a new state-of-the-art. 

\begin{figure*}[t]
  \centering
   \includegraphics[width=1\textwidth]{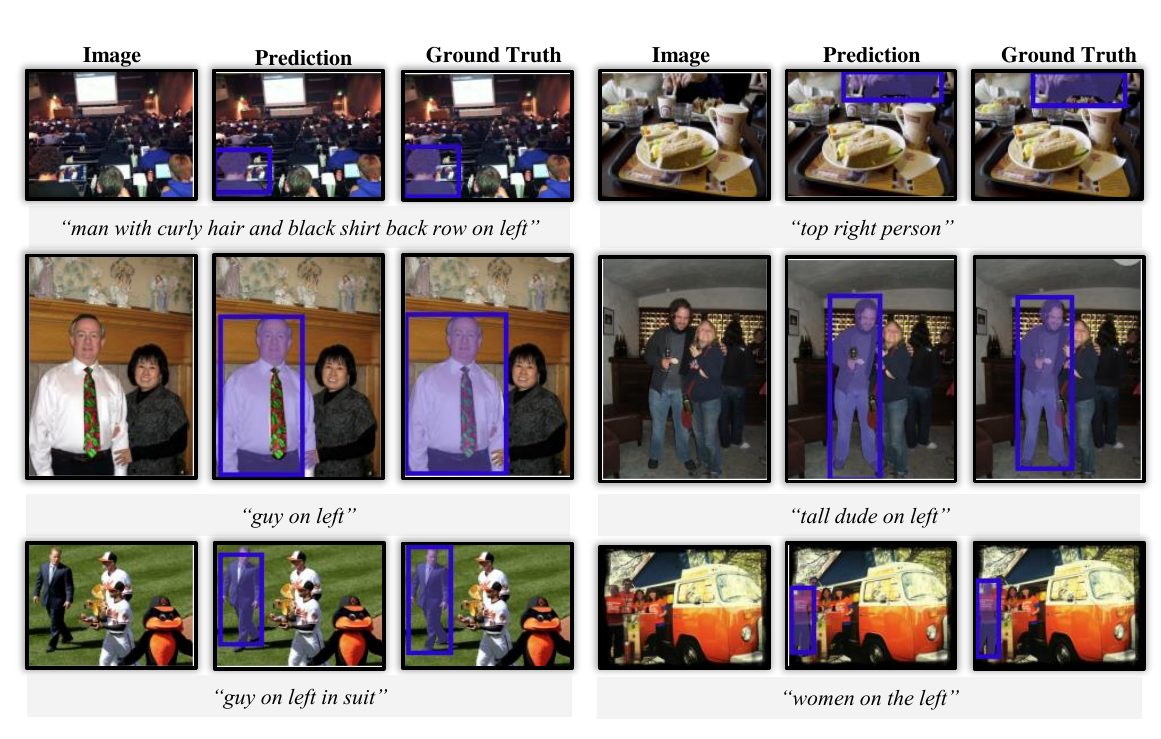}
   \caption{The visualization of zero-shot on REC and RES datsets by NICE.}
   \label{sm_fig6}
\end{figure*}


Compared to RefTR~\citep{li2021referring}, our model demonstrated superior performance across all detection metrics while simultaneously reducing computation needs by four times (808.1 \emph{vs.} 285.2 Gflops), leading to a marked increase in inference speed. Lastly, we compared our model with MCN~\citep{luo2020multi}, which is specifically designed for jointly learning the REC and RES with only one target per image. Despite its purpose-built architecture, MCN could still only generate a mask and a box for each noun phrase one by one when migrating to the PND and PNS tasks. In contrast, our NICE model achieved up to 7.0\% gain on all the masks and 7.8\% on all the boxes, demonstrating its effectiveness in multi-task learning.


In addition, in Fig.~\ref{sm_fig1}, we showcase the distribution of Intersection over Union (IoU) for PNS and PND, where the masks' IoU distribution is displayed in the top row while the IoU distribution for boxes is presented in the lower row. The columns represent all target cases, encompassing both thing and stuff cases as well as singular and plural cases. The noteworthy observation from the results is that our suggested NICE approach achieves the highest recall at all IoU thresholds, making it more efficacious than other methods, as evident from the red curve which consistently outperforms the other curves.

Moreover, we note that the IoU curves for PNS and PND are consistently aligned in NICE, compared to NICE$^\dagger$ and MCN, where the decline trend of the PND and PNS IoU curves varies beyond the 0.5 threshold. This highlights the effectiveness of our designed cascade mechanism, which enables cooperative learning of both tasks, thus ensuring consistent performance throughout.

\subsection{Ablation Studies}

To verify the contribution of each module, we continuously design ablation experiments on the overall framework and each module, respectively.

\textbf{Single-task \emph{vs.} Multi-task.} In Tab.~\ref{table2}, we compare the performance of the collaborative learning framework with the two tasks learned separately. As can be shown, our proposed NICE performs much better than the two tasks learned independently. For OnlyBox, we remove the CGA and BDL and use the kernel feature to predict the box directly. 
As for the OnlyMask, with the reverse guidance of the box, it is also improved by 4.2\% on all the masks. These indicate that under our framework, PND and PNS indeed exhibit a mutually reinforcing effect. 

\begin{figure*}[]
  \centering
   \includegraphics[width=1\textwidth]{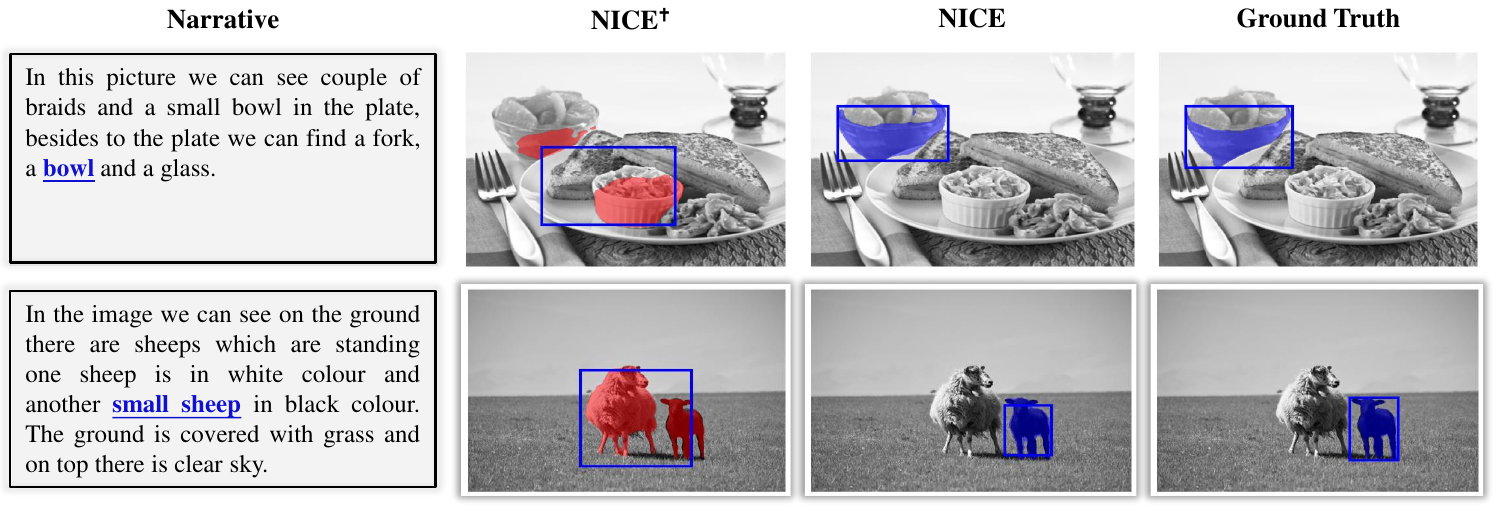}
   \caption{Prediction Conflict, where the wrong parts are marked in red (red boxes of red masks) and the correct parts are marked in blue.}
   \label{sm_fig4}
\end{figure*}

\begin{figure*}[]
  \centering
   \includegraphics[width=1\textwidth]{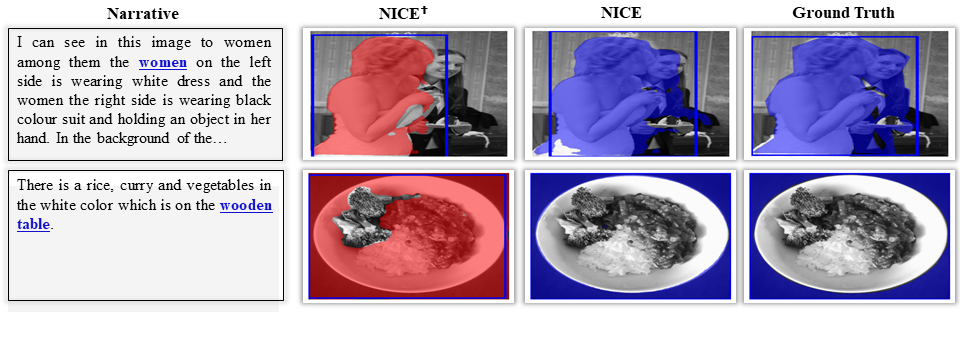}
   \caption{Referring ability comparison between NICE and NICE$^\dagger$}
   \label{sm_fig3}
\end{figure*}

\textbf{With \emph{vs.} without CGA and BDL.} We independently evaluate the performance of CGA and BDL modules in Tab.~\ref{table3}.
The first line indicates that we directly use two task heads in parallel, \emph{i.e.}, NICE$^{\dagger}$. The second line means we use the interacted kernel to directly predict the target instead of the BDL module, which is also a cascade structure. The proposed CGA can bring a 1.8\% improvement on the overall mask. This shows the effectiveness of the cascade structure in promoting the segmentation effect. After adding the BDL module, the detection effect on the overall target is improved by 11.8\%. These results further confirm the merits of BDL in leveraging the barycenters from masks. 

\textbf{The selection of barycenters.} We also conduct ablation experiments on the selection of barycenters in the BDL module in Tab.~\ref{table4}. The \textit{Topk} means we select the $k$ highest-valued point in the mask as barycenters to predict the corresponding boxes, and we will pick the one with the highest confidence as the final predicted box. \textit{Average} is what we mentioned in Sec.~\ref{sec:BDL}. \textit{Average} achieves the best performance. The reason is that the \textit{Average} can best fit the location of the target center. In addition, with the increase in the number of selected points, the recall first increased and then decreased. This is because the likelihood of points close to the center increases as the number of sampled points grows. Yet, when too many points are sampled, too many options can impair the model's convergence.

\textbf{Fused scale \emph{vs.} individual scales.} We discuss the results that the multi-scale scheme brings to BDL. As shown in Tab.~\ref{table5}, We replace the visual features of BDL with multi-scale~\citep{tian2019fcos}. In this case, we find that the detection performance improves by 1.0\% while the segmentation performance decreases by 1.8\%. This is due to the fact that multi-scale information might provide additional candidate boxes to the detection to fit targets at various scales. However, for pixel-level interaction segmentation, features at different scales can cause some confusion.

\textbf{With \emph{vs.} without stuff.} PND is distinct from visual-only detection as it emphasizes vision-language alignment where all nouns, including things and stuff, are significant. To assess the impact of stuff detection, we conducted an ablation study, and the findings in Tab.~\ref{tab: stuff} indicate that removing stuff detection resulted in a decline in performance across all metrics. Notably, the loss in detection accuracy was three times greater than that of segmentation. These results highlight the crucial role of the rich semantics found in stuff for multi-modal grounding.

\textbf{Different approach for bounding boxes.} The NICE$^{\dagger}$ in Tab.~\ref{table1} presents a solution to two parallel task branches. By employing a cascade structure of segmentation followed by detection, we achieve a significant performance boost with only a minor increase in inference time, particularly for the PND task. These results serve as a testament to the effectiveness of the NICE$^{\dagger}$ approach. In addition to NICE$^{\dagger}$, we also introduce a robust reference model, NICE$^{*}$. This model can identify the corresponding box directly by analyzing the shape of the mask. However, NICE$^{*}$ falls short in detection performance compared to NICE. This is because NICE$^{*}$ relies heavily on segmented edges, which can be challenging to manage and ultimately limit its detection accuracy.

IE represents the Inconsistency Error metric~\citep{luo2020multi} to evaluate the prediction conflict between the segmentation and detection task. Our cascade paradigm greatly reduces the prediction conflict compared to the two-branch paradigm, even lower than MCN which has additional alignment and post-processing operations. This validates the ability of NICE to infer in collaboration across tasks.

\subsection{Expand the Boundary of NICE}

\subsubsection{Train NICE with Only Bounding Boxes}
To verify the robustness of NICE on more benchmarks, we conduct the experiment on Flickr30k~\citep{plummer2015flickr30k}, focusing on detect multi-objects with the referred texts. NICE can be supervised merely by bounding boxes, without the necessity of segmentation labels. This makes NICE versatile, not limited to PNG but extendable to other Visual Grounding benchmarks. In this setup, segmentation module acts akin to an attention mechanism, delivering crucial location and contour details to aid detection. with the results tabulated in Tab.~\ref{tab:flickr30k}. Our NICE outstrips the non-pre-trained state-of-the-art (FAOA) and simultaneously increases inference speed by more than threefold. 

\subsubsection{Zero-Shot Study for RES and REC}
In order to evaluate NICE's ability in zero-shot scenarios, we conducted experiments on the REC and RES datasets, and achieved significant improvement in AP@0.5, as shown in Tab.~\ref{tablezs} and Tab.~\ref{tablezb}. During training, the full phrase was used as the text feature, which further demonstrated NICE's exceptional generalizability, remaining stable even when tested on the challenging dataset, \emph{i.e.,} RefCOCO+. Our findings suggest that our framework presents a highly versatile paradigm for visual grounding tasks.

To examine NICE's ability to capture the connection between image and referred text, we investigated zero-shot settings using visual examples, presented in Fig.~\ref{sm_fig6}. Impressively, NICE generated more precise masks than the ground truth, as exemplified in the second row of Fig.~\ref{sm_fig6}, which resulted from the highly refined semantic alignment obtained via the collaborative learning brought on by the PND and PNS tasks. It is notable that NICE was able to differentiate the person, tie, and microphone with impressive accuracy.

\subsection{Qualitative Analysis}
\subsubsection{visualizations}
As shown in Fig.~\ref{fig4}, we present some typical grounding results from NICE compared to the ground truth. In fact, as we expected, the segmentation results of the NICE are accurate, showing the correct pointing ability, which means that the model accurately understood both the semantic and image information. To our surprise, the performance of the model is even better than the ground truth in some areas. We have framed these parts in ground truth with white boxes. For the first image, the ground truth incorrectly labels the air conditioner as a wall. In the second image as well, the ground truth does not split the beach, water, and hills as accurately as NICE does.

In addition, we compare the NICE with NICE$^\dagger$ to prove the validity of CGA and BDL in Fig.~\ref{fig5}. As we can see, NICE handles well the problem of conflicting predictions that NICE$^\dagger$ cannot resolve.

\subsubsection{Prediction Conflict}
We also supply more visualizations to explore predicted conflict phenomena in NICE and NICE$^\dagger$, as shown in Fig.~\ref{sm_fig4}. We mark the wrong part in red (red boxes of red masks) and the correct part in blue. 
In NICE$^\dagger$, box variations or size inconsistencies are common in detection errors, and the case of segmentation error is primarily concerned with the scenario of an incomplete mask. This is because the two-branch structure is too independent and lacks synergy. In contrast, this phenomenon is greatly reduced in NICE with cascade structure. This is because we leverage the barycenter of the segmentation mask as an anchor to connect segmentation and detection in series so that the two tasks are naturally aligned with each other.

\subsubsection{Referring Ability}
As for Fig.~\ref{sm_fig3}, we particularly analyze the superior referential ability brought by BDL in PNS. In the NICE$^\dagger$ model, PNS falls short in discriminating among visually similar targets, particularly when the objects in question are indistinguishable. This issue arises from its reliance on kernel-to-feature similarity computations for obtaining response pixels. Addressing this, our proposal for NICE presents a solution by employing a BDL approach that leverages spatial information to pinpoint similar targets appearing in different areas. This effectively circumvents the problem as shown in Fig.~\ref{sm_fig3}.

\section{Conclusion}

In this paper, we proposed a unified cascading framework termed NICE for panoptic narrative detection and segmentation. To better handle the prediction conflict problem of PNS and PND, two innovative designs are proposed based on the barycenter of segmentation mask, namely Coordinate Guided Aggregation (CGA) and Barycenter Driven Localization (BDL), respectively. Extensive experiments show that our proposed NICE achieves SOTA compared to the existing SOTA methods with faster inference (3x). Furthermore, the proposed BDL is a novel detection paradigm that inherits unified thought and eliminates standard object detection's reliance on a large number of pre-selected anchors. We believe this might be expanded to the typical object detection area, increasing detection inference speed while enhancing interpretability.

%

\begin{appendices}




\end{appendices}


\bibliographystyle{bst/sn-basic}
\bibliography{sn-bibliography} 


\end{document}